\def\year{2018}\relax
\documentclass[letterpaper]{article} %DO NOT CHANGE THIS
\usepackage{aaai18}  %Required
\usepackage{times}  %Required
\usepackage{helvet}  %Required
\usepackage{courier}  %Required
\usepackage{url}  %Required
\usepackage{graphicx}  %Required
\usepackage{times}
\usepackage{latexsym}

\usepackage{caption}
\usepackage{xcolor}
\usepackage{booktabs}
\usepackage{multirow}

\usepackage{wrapfig}
\usepackage{hyperref}
\usepackage[font=small,labelfont=bf]{caption}

\newcommand{\citep}{\cite}
\newcommand{\citet}[1]{\citeauthor{#1} (\citeyear{#1})}

\newif\ifacl

\acltrue

\begin{document}
\title{Gated-Attention Architectures for Task-Oriented Language Grounding}
\author{Devendra Singh Chaplot\\
\texttt{chaplot@cs.cmu.edu}\\
School of Computer Science\\
Carnegie Mellon University  
\And
Kanthashree Mysore Sathyendra\thanks{ Equal contribution} \\
\texttt{ksathyen@cs.cmu.edu}\\
School of Computer Science\\
Carnegie Mellon University 
\AND
Rama Kumar Pasumarthi\footnotemark[1] \\
\texttt{rpasumar@cs.cmu.edu}\\
School of Computer Science\\
Carnegie Mellon University 
\And
Dheeraj Rajagopal\footnotemark[1] \\
\texttt{dheeraj@cs.cmu.edu}\\
School of Computer Science\\
Carnegie Mellon University 
\And
Ruslan Salakhutdinov\\
\texttt{rsalakhu@cs.cmu.edu}\\
School of Computer Science\\
Carnegie Mellon University  
 }
\maketitle
\begin{abstract}
To perform tasks specified by natural language instructions, autonomous agents need to extract semantically meaningful representations of language and map it to visual elements and actions in the environment. This problem is called task-oriented language grounding. We propose an end-to-end trainable neural architecture for task-oriented language grounding in 3D environments which assumes no prior linguistic or perceptual knowledge and requires only raw pixels from the environment and the natural language instruction as input. The proposed model combines the image and text representations using a Gated-Attention mechanism and learns a policy to execute the natural language instruction using standard reinforcement and imitation learning methods. We show the effectiveness of the proposed model on unseen instructions as well as unseen maps, both quantitatively and qualitatively. We also introduce a novel environment based on a 3D game engine to simulate the challenges of task-oriented language grounding over a rich set of instructions and environment states.
\end{abstract}

\vspace{-0.2in}
\section{Introduction} 
%\vspace{-0.05in}

Artificial Intelligence (AI) systems are expected to perceive the environment and take actions to perform a certain task \citep{russell1995modern}. 
Task-oriented language grounding refers to the process of extracting semantically meaningful representations of language by mapping it to visual elements and actions in the environment in order to perform the task specified by the instruction.

\begin{figure}
\centering
\includegraphics[width=0.99\linewidth,keepaspectratio]{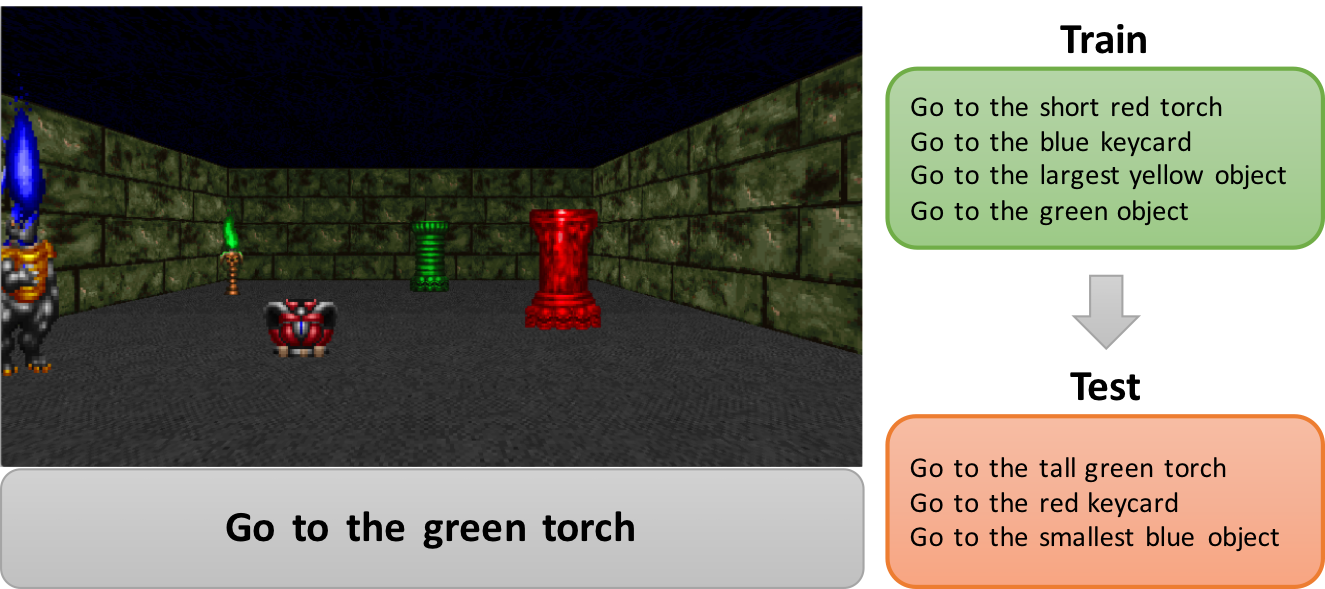}
%%\vspace{-10pt}
\caption{\small An example of task-oriented language grounding in the 3D Doom environment with sample instructions. The test set consists of unseen instructions.}
\label{fig:scrn}
\vspace{9pt}
\end{figure}

Consider the scenario shown in Figure \ref{fig:scrn}, where an agent takes natural language instruction and pixel-level visual information as input to carry out the task in the real world. To accomplish this goal, the agent has to draw semantic correspondences between the visual and verbal modalities and learn a policy to perform the task. This problem poses several challenges: the agent has to learn to \emph{recognize} objects in raw pixel input, \emph{explore} the environment as the objects might be occluded or outside the field-of-view of the agent, \emph{ground} each concept of the instruction in visual elements or actions in the environment, \emph{reason} about the pragmatics of language based on the objects in the current environment (for example instructions with superlative tokens, such as `Go to the largest object') and \emph{navigate} to the correct object while avoiding incorrect ones.

\begin{figure*}
\centering
\includegraphics[width=0.99\linewidth,keepaspectratio]{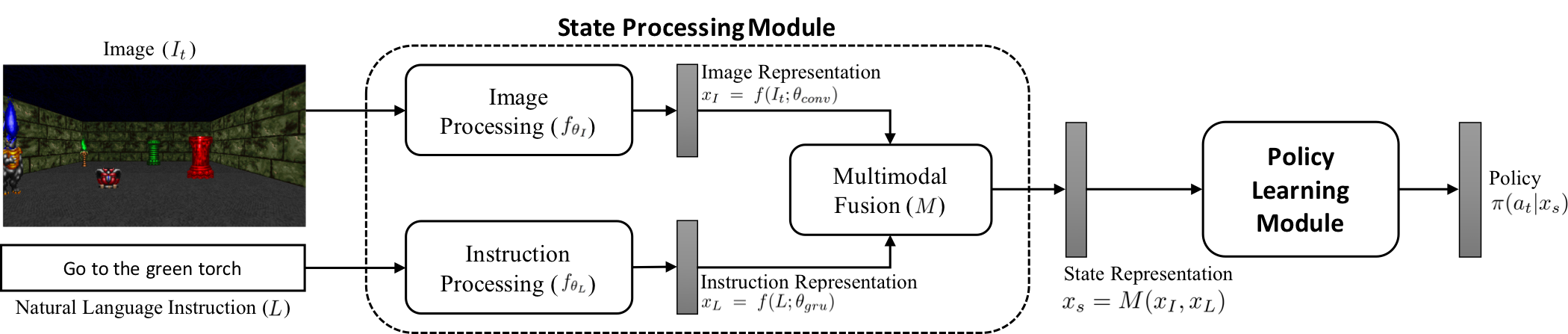}
%\vspace{-5pt}
\caption{\small The proposed model architecture to estimate the policy given the natural language instruction and the image showing the first-person view of the environment.}
\label{fig:arch}
%\vspace{-12pt}
\end{figure*}

To tackle this problem, we propose an architecture that comprises of a \textit{state processing module} that creates a joint representation of the instruction and the images observed by the agent, and a \textit{policy learner} to predict the optimal action the agent has to take in that timestep.   
The state processing module consists of a novel Gated-Attention multimodal fusion mechanism, 
which is based on multiplicative interactions between both modalities~\citep{dhingra2016gated,Wu2016}. 
 
The contributions of this paper are summarized as follows:
1) We propose an end-to-end trainable architecture that handles raw pixel-based input for task-oriented language grounding in a 3D environment and assumes no prior linguistic or perceptual knowledge\footnote{The environment and code is available at \url{https://github.com/devendrachaplot/DeepRL-Grounding}}. We show that the proposed model generalizes well to unseen instructions as well as unseen maps\footnote{See demo videos at \url{https://goo.gl/rPWlMy}}.
2) We develop a novel Gated-Attention mechanism for multimodal fusion of representations of verbal and visual modalities. We show that the gated-attention mechanism outperforms the baseline method of concatenating the representations using various policy learning methods. The visualization of the attention weights in the gated-attention unit shows that the model learns to associate attributes of the object mentioned in the instruction with the visual representations learned by the model.
3) We introduce a new environment, built over ViZDoom \citep{kempka2016vizdoom}, for task-oriented language grounding with a rich set of actions, objects and their attributes. The environment provides a first-person view of the world state, and allows for simulating complex scenarios for tasks such as navigation.
%\footnote{The code for the environment and the proposed model is available at \url{https://github.com/devendrachaplot/DeepRL-Grounding}}

\section{Related Work}
\label{sec:rel}
%\vspace{-0.05in}

\textbf{Grounding Language in Robotics}. In the context of grounding language in objects and their attributes, \citet{guadarrama2014open} present a method to ground open vocabulary to objects in the environment. Several works look at grounding concepts through human-robot interaction \citep{chao2011towards,lemaignan2012grounding}. Other works in grounding include attempts to ground natural language instructions in haptic signals \citep{chu2013using} and teaching robot to ground natural language using active learning \citep{kulick2013active}. 
Some of the work that aims to ground navigational instructions include \citep{guadarrama2013grounding}, \citep{bollini2013interpreting} and \citep{beetz2011robotic}, where the focus was to ground verbs like \emph{go, follow, etc.} and spatial relations of verbs \citep{tellex2011understanding,fasola2013using}. 

\textbf{Mapping Instructions to Action Sequences}.
\citet{chen2011learning} and \citet{artzi2013weakly} present methods based on semantic parsing to map navigational instructions to a sequence of actions. \citet{mei2015listen} look at neural mapping of instructions to sequence of actions, along with input from bag-of-word features extracted from the visual image. While these works focus on grounding navigational instructions to actions in the environment, we aim to ground visual attributes of objects such as shape, size and color.  

\textbf{Deep reinforcement learning using visual data.} Prior work has explored using Deep Reinforcement learning approaches for playing FPS games \citep{lample2016playing,kempka2016vizdoom,kulkarni2016deep}. The challenge here is to learn optimal policy for a variety of tasks, including navigation using raw visual pixel information. \citet{chaplottransfer} look at transfer learning between different tasks in the Doom Environment. In all these methods, the policy for each task is learned separately using a deep Q-Learning \citep{mnih2013playing}. In contrast, we train a single network for multiple tasks/instructions. 
\citet{zhu2016target} look at target-driven visual navigation, given the image of the target object. We use the natural language instruction and do not have the visual image of the object.
\citet{yu2017deep} look at learning to navigate in a 2D maze-like environment and execute commands, for both seen and \textit{zero-shot} setting, where the combination of words are not seen before. \citet{misra2017mapping} also look at mapping raw visual observations and text input to actions in a 2D Blocks environment. 
While these works also looks at executing a variety of instructions, they tackle only 2D environments. \citet{oh2017zero} look at zero-shot task generalization in a 3D environment. Their method tackles long instructions with several subtasks and a wide variety of action verbs. However, the position of the agent is discretized like 2D Mazes and their method encodes some prior linguistic knowledge in a analogy making objective.

Compared to the prior work, this paper aims to address visual language grounding in a challenging 3D setting involving raw-pixel input, continuous agent positions and partially observable envrionment, which poses additional challenges of perception, exploration and reasoning. Unlike many of the previous methods, our model assumes no prior linguistic or perceptual knowledge, and is trainable end-to-end.

\section{Problem Formulation}
\label{sec:prob}
%\vspace{-0.05in}

We tackle the problem of task-oriented language grounding in the context of target-driven visual navigation conditioned on a natural language instruction, where the agent has to navigate to the object described in the instruction. 
Consider an agent interacting with an episodic environment~$\mathcal{E}$. 
In the beginning of each episode, the agent receives a natural language instruction ($L$) which indicates the description of the target, a visual object in the environment. 
At each time step, the agent receives a raw pixel-level image of the first person view of the environment ($I_t$), and performs an action $a_t$. 
The episode terminates whenever the agent reaches any object or the number of time steps exceeds the maximum episode length. 
Let $s_t = \{I_t,L\}$ denote the state at each time step. 
The objective of the agent is to learn an optimal policy $\pi(a_t|s_t)$, which maps the observed states to actions, eventually leading to successful completion of the task. In this case, the task is to reach the correct object before the episode terminates. 
We consider two different learning approaches:
(1) Imitation Learning \citep{bagnell2015invitation}: where the agent has access to an oracle which specifies the optimal action given any state in the environment;
(2) Reinforcement Learning \citep{sutton1998reinforcement}: where the agent receives a positive reward when it reaches the target object and a negative reward when it reaches any other object.

\section{Proposed Approach}
\label{sec:model-arch}
%\vspace{-0.05in}

We propose a novel architecture for task-oriented visual language grounding, which assumes no prior linguistic or perceptual knowledge and can be trained end-to-end. The proposed model is divided into two modules, state processing and policy learning, as shown in Figure~\ref{fig:arch}. 

{\bf State Processing Module}: The state processing module takes the current state $s_t = \{I_t,L\}$ as the input and creates a joint representation for the image and the instruction. 
This joint representation is used by the policy learner to predict the optimal action to take at that timestep. 
It consists of a convolutional network \citep{lecun1995convolutional} to process the image $I_t$, a Gated Recurrent Unit (GRU) \citep{cho2014properties} network to process the instruction~$L$ and
a multimodal fusion unit that combines the representations of the instruction and the image. Let $x_I = f(I_t;\theta_{conv}) \in \mathcal{R}^{d\times H\times W}$ be the representation of the image, where $\theta_{conv}$ denote the parameters of the convolutional network, $d$ denotes number of feature maps (intermediate representations) in the convolutional network output, while $H$ and $W$ denote the height and width of each feature map. Let $x_L = f(L;\theta_{gru})$ be the representation of the instruction, where $\theta_{gru}$ denotes the parameters of the GRU network. The multimodal fusion unit, $M(x_I,x_L)$ combines the image and instruction representations. Many prior methods combine the multimodal representations by concatenation \cite{mei2015listen,misra2017mapping}. We develop a multimodal fusion unit, \textit{Gated-Attention}, based on multiplicative interactions between instruction and image representation.

%\vspace{3pt}
{\bf Concatenation}:
In this approach, the representations of the image and instruction are simply flattened and concatenated to create a joint state representation:
%\vspace{-3pt}
$$ M_{concat}(x_I,x_L) =  [\textrm{vec}(x_I);\textrm{vec}(x_L)],$$ 

%\vspace{-3pt}
\noindent where $\textrm{vec}(.)$ denotes the flattening operation. The concatenation unit is used as a baseline for the proposed Gated-Attention unit as it is used by prior methods \cite{mei2015listen,misra2017mapping}.

\begin{figure}
\includegraphics[width=0.95\linewidth,height=\textheight,keepaspectratio]{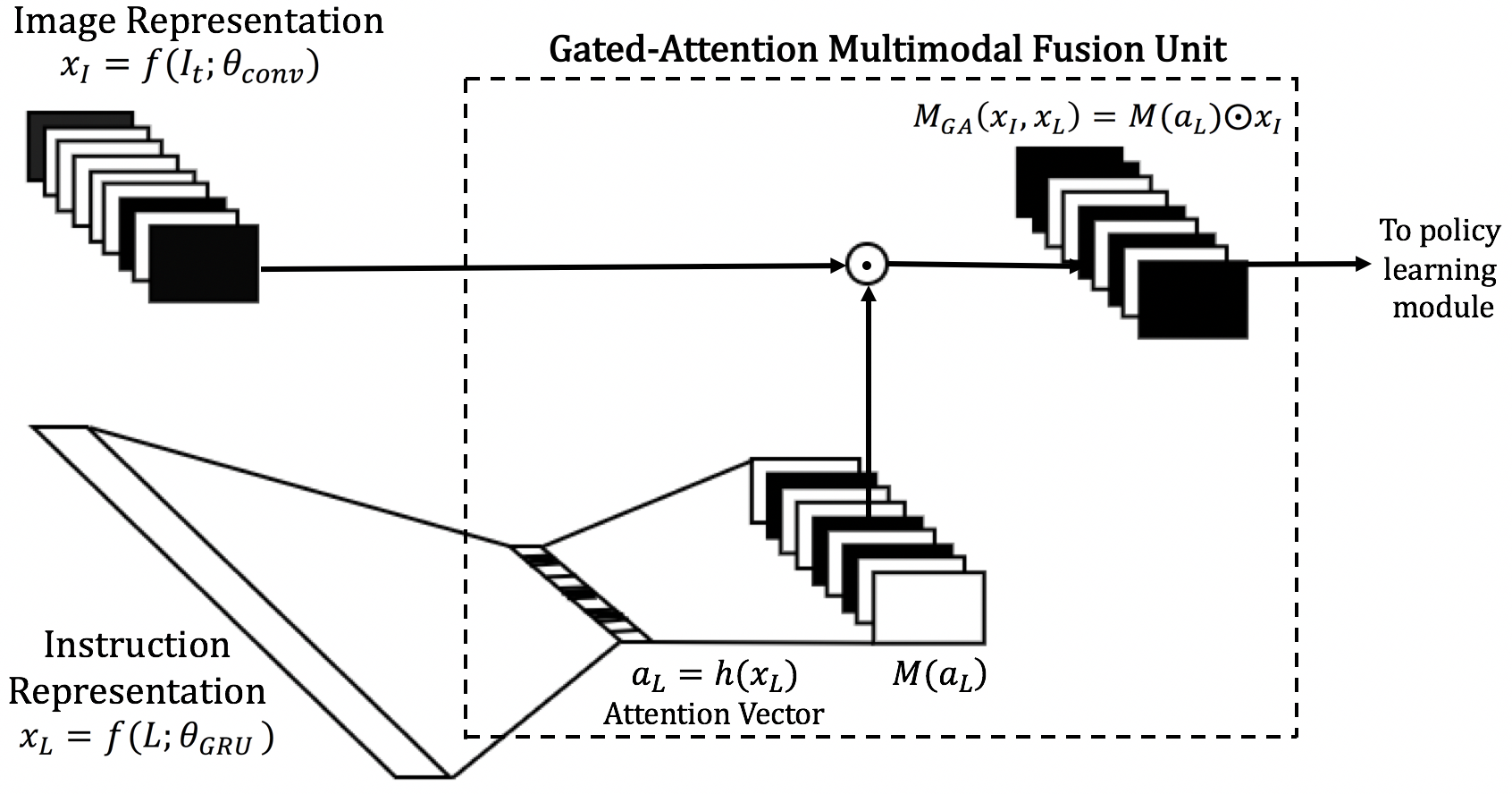}
%\vspace{-15pt}
\caption{Gated-Attention unit architecture.}
\label{fig:ga}
%\vspace{-12pt}
\end{figure}

\begin{figure}
\includegraphics[width=0.95\linewidth,height=\textheight,keepaspectratio]{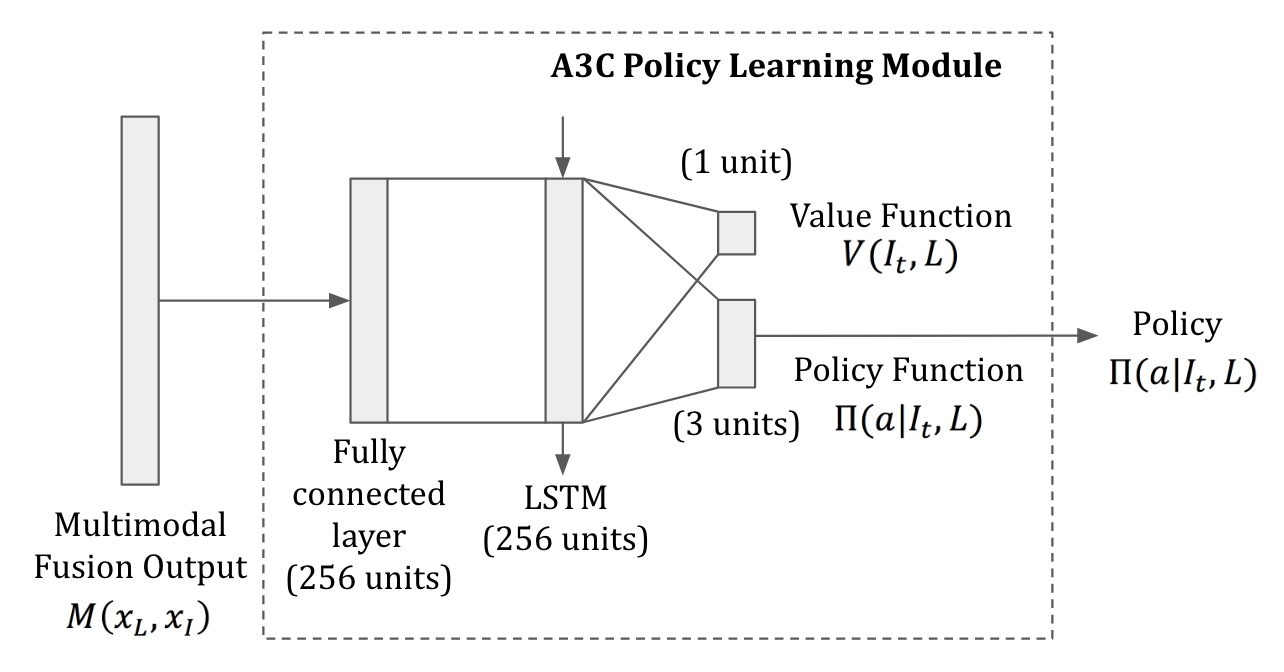}
%\vspace{-15pt}
\caption{A3C policy model architecture.}
\label{fig:a3c}
%\vspace{-12pt}
\end{figure}

%\vspace{3pt}
{\bf Gated-Attention}:
In the Gated-Attention unit, the instruction embedding is passed through a fully-connected linear layer with a sigmoid activation. The output dimension of this linear layer, $d$, is equal to the number of feature maps in the output of the convolutional network (first dimension of $x_I$). The output of this linear layer is called the attention vector $a_L = h(x_L) \in \mathcal{R}^d$, where $h(.)$ denotes the fully-connected layer with sigmoid activation. Each element of $a_L$ is expanded to a $H \times W$ matrix. This results in a  3-dimensional matrix,  $M({a_L}) \in \mathcal{R}^{d\times H \times W}$  whose $(i,j,k)^{th}$ element is given by: $M_{a_L}[i,j,k] = a_L[i]$.
This matrix is multiplied element-wise with the output of the convolutional network:
%\vspace{-3pt}
$$M_{GA}(x_I,x_L) = M(h(x_L)) \odot x_I = M(a_L)\odot x_I,$$

%\vspace{-3pt}
\noindent where $\odot$ denotes the Hadamard product \citep{horn1990hadamard}. The architecture of the Gated-Attention unit is shown in Figure~\ref{fig:ga}. The whole unit is differentiable which makes the architecture end-to-end trainable. 

The proposed Gated-Attention unit is inspired by the Gated-Attention Reader architecture for text comprehension \citep{dhingra2016gated}. They integrate a multi-hop architecture with a Gated-attention mechanism, which is based on multiplicative interactions between the query embedding and the intermediate states of a recurrent neural network document reader. In contrast, we propose a Gated-Attention multimodal fusion unit which is based on multiplicative interactions between the instruction representation and the convolutional feature maps of the image representation. This architecture can be extended to any application of multimodal fusion of verbal and visual modalities.

The intuition behind Gated-Attention unit is that the trained convolutional feature maps detect different attributes of the objects in the frame, such as color and shape. The agent needs to attend to specific attributes of the objects based on the instruction. For example, depending on the whether the instruction is ``Go to the green object'', ``Go to the pillar'' or ``Go to the green pillar'' the agent needs to attend to objects which are `green', objects which look like a `pillar' or both. The Gated-Attention unit is designed to gate specific feature maps based on the attention vector from the instruction, $a_L$. 

\subsection{Policy Learning Module}
%\vspace{-0.05in}
The output of the multimodal fusion unit ($M_{concat}$ or $M_{GA}$) is fed to the policy learning module. The architecture of the policy learning module is specific to the learning paradigm: (1) Imitation Learning or (2) Reinforcement Learning.

For imitation learning, we consider two algorithms, Behavioral Cloning \citep{bagnell2015invitation} and DAgger \citep{ross2011reduction}. Both the algorithms require an oracle that can return an optimal action given the current state. The oracle is implemented by extracting agent and target object locations and orientations from the Doom game engine. Given any state, the oracle determines the optimal action as follows: The agent first reorients (using \textit{turn\_left, turn\_right} actions) towards the target object. It moves forward (\textit{move\_forward} action), reorienting towards the target object if deviation of the agent's orientation is greater than the minimum turn angle supported by the environment. 

For reinforcement learning, we use the Asynchronous Advantage Actor-Critic (A3C) algorithm \citep{mnih2016asynchronous} which uses a deep neural network to learn the policy and value functions and runs multiple parallel threads to update the network parameters. We also use the entropy regularization for improved exploration as described by \citep{mnih2016asynchronous}. 
In addition, we use the Generalized Advantage Estimator \citep{schulman2015high} to reduce the variance of the policy gradient \citep{williams1992simple} updates. 

The policy learning module for imitation learning contains a fully connected layer to estimate the policy function.
The policy learning module for reinforcement learning using A3C (shown in Figure \ref{fig:a3c}) consists of an LSTM layer, followed by fully connected layers to estimate the policy function as well as the value function. The LSTM layer is introduced so that the agent can have some memory of previous states. This is important as a reinforcement learning agent might explore states where all objects are not visible and need to remember the objects seen previously.

\begin{figure}
\centering
\includegraphics[width=0.85\linewidth,height=\textheight,keepaspectratio]{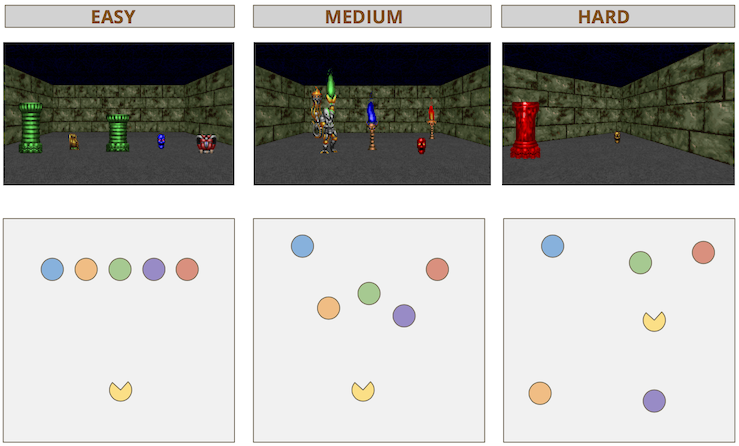}
%\vspace{-5pt}
\caption{Sample starting states and bird's eye view of the map (not visible to the agent) showing agent and object locations in Easy, Medium and Hard modes.}
\label{fig:env_sample}
%\vspace{-12pt}
\end{figure}

\section{Environment}
\label{sec:environment}
%\vspace{-0.05in}

We create an environment for task-oriented language grounding, where the agent can execute a natural language instruction and obtain a positive reward on successful completion of the task. 
Our environment is built on top of the ViZDoom API \citep{kempka2016vizdoom}, based on Doom, a classic first person shooting game. It provides the raw visual information from a first-person perspective at every timestep. Each scenario in the environment comprises of an agent and a list of objects (a subset of ViZDoom objects) - one correct and rest incorrect in a customized map. The agent can interact with the environment by performing navigational actions such as turn left, turn right, move forward. Given an instruction ``Go to the green torch", the task is considered successful if the agent is able to reach the \emph{green torch} correctly. 
The customizable nature of the environment enables us to create scenarios with varying levels of difficulty which we believe leads to designing sophisticated learning algorithms to address the challenge of multi-task and zero-shot reinforcement learning. 

An \emph{instruction} is a combination of (action, attribute(s), object) triple. Each instruction can have more than one attribute but we limit the number of actions and objects to one each.  
The environment  allows a variety of objects to be spawned at different locations in the map. The objects can have various visual attributes such as color, shape and size\footnote{See Appendix for the list of objects and instructions}. We provide a set of 70 manually generated instructions\footnotemark[3]. For each of these instructions, the environment allows for automatic creation of multiple episodes, each randomly created with its own set of correct object and incorrect objects.
Although the number of instructions are limited, the combinations of correct and incorrect objects for each instruction allows us to create multiple settings for the same instruction. Each time an instruction is selected, the environment generates a random combination of incorrect objects and the correct object in randomized locations. 
One of the significant challenges posed for a learning algorithm is to understand that the same instruction can refer to different objects in the different episodes. For example, ``Go to the red object'' can refer to a red keycard in one episode, and a red torch in another episode. Similarly, ``Go to the keycard'' can refer to keycards of various colors in different episodes. 
Objects could also occlude each other, or might not even be present in the agent's field of view, or the map could be more complicated, making it difficult for the agent to make a decision based solely on the current input, stressing the need for efficient exploration.

Our environment also provides different modes with respect to spawning of objects each with varying difficulty levels (Figure~\ref{fig:env_sample}):
\textbf{Easy}: The agent is spawned at a fixed location. The candidate objects are spawned at five fixed locations along a single horizontal line along the field of view of the agent.
\textbf{Medium}: The candidate objects are spawned in random locations, but the environment ensures that they are in the field of view of the agent. The agent is still spawned at a fixed location.
\textbf{Hard}: The candidate objects and the agent are spawned at random locations and the objects may or may not be in the agents field of view in the initial configuration. The agent needs to explore the map to view all the objects.

\begin{figure*}
\centering
\begin{minipage}{0.32\textwidth}
\includegraphics[width=\linewidth,height=\textheight,keepaspectratio]{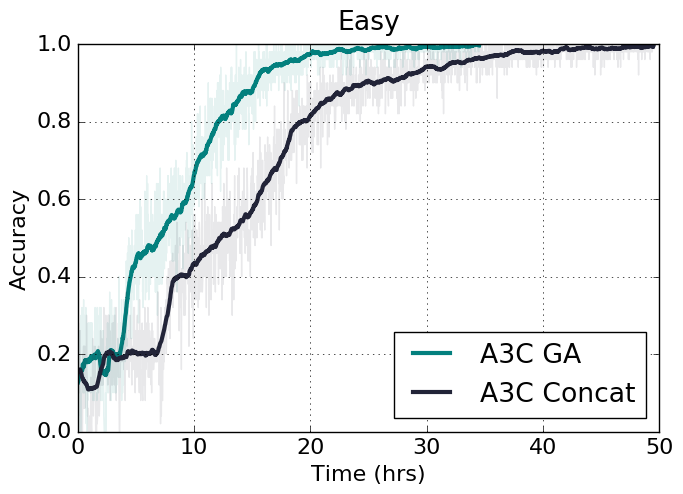}
\end{minipage}
\begin{minipage}{0.32\textwidth}
\centering
\includegraphics[width=\linewidth,height=\textheight,keepaspectratio]{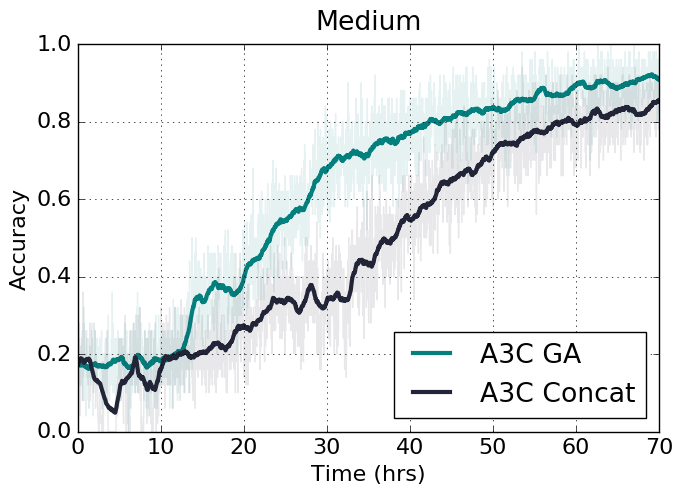}
\end{minipage}
\begin{minipage}{0.32\textwidth}
\centering
\includegraphics[width=\linewidth,height=\textheight,keepaspectratio]{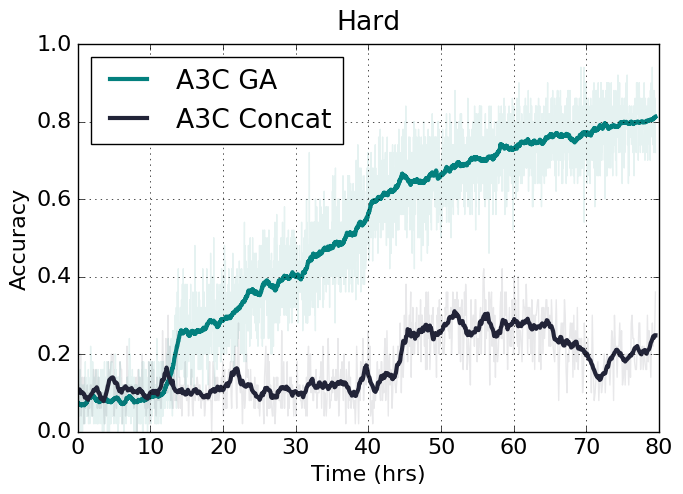}
\end{minipage}
%\vspace{-5pt}
\caption{\small Comparison of the performance of the proposed Gated-Attention (GA) unit to the baseline Concatenation unit using Reinforcement learning algorithm, A3C for (a) easy, (b) medium and (c) hard environments.}
\label{fig:policy_evaluation}
%\vspace{-5pt}
\end{figure*}

\section{Experimental Setup}
\label{sec:exp}
%\vspace{-0.05in}
We perform our experiments
%\footnote{The code for the environment and the proposed model is available at \url{https://github.com/devendrachaplot/DeepRL-Grounding}} 
in all of the three environment difficulty modes, where we restrict the number of objects to 5 for each episode (one correct object, four incorrect objects and the agent are spawned for each episode). 
During training, the objects are spawned from a training set of 55 instructions, while 15 instructions pertaining to unseen \textit{attribute-object} combinations are held out in a test set for zero-shot evaluation.
During training, at the start of each episode, one of the train instructions is selected randomly. A correct target object is selected and 4 incorrect objects are selected at random. These objects are placed at random locations depending on the difficulty level of the environment. The episode terminates if the agent reaches any object or time step exceeds the maximum episode length ($T=30$). The evaluation metric is the \emph{accuracy} of the agent which is success rate of reaching the correct object before the episode terminates. 
We consider two scenarios for evaluation:

\noindent (1) \textbf{Multitask Generalization} (MT), where the agent is evaluated on unseen maps with instructions in the train set. Unseen maps comprise of unseen combination of objects placed at randomized locations. This scenario tests that the agent doesn't overfit to or memorize the training maps and can execute multiple instructions or tasks in unseen maps. 

\noindent (2) \textbf{Zero-shot Task Generalization} (ZSL), where the agent is evaluated on unseen test instructions. This scenario tests whether the agent can generalize to new combinations of attribute-object pairs which are not seen during the training. The maps in this scenario are also unseen.

\subsection{Baseline Approaches}
\textbf{Reinforcement Learning:} We adapt \citep{misra2017mapping} as a reinforcement learning baseline in the proposed environment. \citet{misra2017mapping} looks at jointly reasoning on linguistic and visual inputs for moving blocks in a 2D grid environment to execute an instruction. Their work uses raw features from the 2D grid, processed by a CNN, while the instruction is processed by an LSTM. Text and visual representations are combined using concatenation. The agent is trained using reinforcement learning and enhanced using distance based reward shaping. We do not use reward shaping as we would like the method to generalize to environments where the distance from the target is not available. 

\noindent \textbf{Imitation Learning:} We adapt \citep{mei2015listen} as an imitation learning baseline in the proposed environment. \citet{mei2015listen} map sequence of instructions to actions, treated as a sequence-to-sequence learning problem, with visual state input received by the decoder at each decode timestep. While they use a bag-of-visual words representation for visual state, we adapt the baseline to directly process raw pixels from the 3D environment using CNNs.

To ensure fairness in comparison, we use exact same architecture of CNNs (to process visual input), GRUs (to process textual instruction) and policy learning across baseline and proposed models. This reduces the reinforcement learning baseline to A3C algorithm with concatenation multimodal fusion (\textit{A3C-Concat}), and imitation learning baseline to Behavioral Cloning with Concatenation (\textit{BC-Concat}).

\setlength{\tabcolsep}{1.2em}
\begin{table*}[]
\small  
\centering
\begin{tabular}{@{}clcllllll@{}}
\toprule
\multicolumn{2}{c}{\multirow{2}{*}{Model}}                                                        & \multicolumn{1}{c}{\begin{tabular}[c]{@{}c@{}}Parameters\end{tabular}} & \multicolumn{2}{c}{Easy} & \multicolumn{2}{c}{Medium} & \multicolumn{2}{c}{Hard} \\
\multicolumn{2}{c}{}                                                                              &                                                                                     & MT          & ZSL        & MT          & ZSL          & MT          & ZSL        \\ \midrule
\multirow{4}{*}{\begin{tabular}[c]{@{}c@{}}Imitation\\ Learning\end{tabular}}                                            
& BC Concat     & 5.21M & 0.86 & 0.71 & 0.23 & 0.15 & 0.20 & 0.15 \\
& BC GA         & 5.09M & \bf{0.97} & 0.81 & 0.30 & 0.23 & \bf{0.36} & 0.29 \\
& DAgger Concat & 5.21M & 0.92 & 0.73 & 0.45 & 0.23 & 0.19 & 0.13 \\
& DAgger GA     & 5.09M & 0.94 & \bf{0.85} & \bf{0.55} & \bf{0.40} & 0.29 & \bf{0.30} \\ \midrule
\multirow{2}{*}{\begin{tabular}[c]{@{}c@{}}Reinforcement\\ Learning\end{tabular}} 
& A3C Concat    & 3.44M                                                                               & 1.00        & 0.80           & 0.80            & 0.54             & 0.24            &  0.12         \\
& A3C GA        & 3.39M                                                                               & 1.00        & \bf{0.81}       & \bf{0.89}            & \bf{0.75}             & \bf{0.83}            & \bf{0.73}           \\ \bottomrule
\end{tabular}%
\vspace{-1pt}
\caption{The accuracy of all the models with Concatenation and Gated-Attention (GA) units. A3C Concat and BC Concat are the adapted versions of \protect\citet{misra2017mapping}  and \protect\citet{mei2015listen} respectively for the proposed environment. All the accuracy values are averaged over 100 episodes.}
\label{tab:results}
%\vspace{-5pt}
\end{table*}

%\vspace{-0.05in}
\subsection{Hyper-parameters}
%\vspace{-0.05in}
The input to the neural network is the instruction and an RGB image of size 3x300x168. The first layer convolves the image with 128 filters of 8x8 kernel size with stride 4, followed by 64 filters of 4x4 kernel size with stride 2 and another 64 filters of 4x4 kernel size with stride 2. The architecture of the convolutional layers is adapted from previous work on playing deathmatches in Doom \cite{chaplot2017arnold}. The input instruction is encoded through a Gated Recurrent Unit (GRU) \citep{chung2014empirical} of size 256. 

For the imitation learning approach, we run experiments with Behavioral Cloning (BC) and DAgger algorithms in an online fashion, which have data generation and policy update function per outer iteration. The policy learner for imitation learning comprises of a linear layer of size 512 which is fully-connected to 3 neurons to predict the policy function (i.e. probability of each action). In each data generation step, we sample state trajectories based on oracle's policy in BC and based on a mixture of oracle's policy and the currently learned policy in DAgger. The mixing of the policies is governed by an exploration coefficient, which has a linear decay from 1 to 0. For each state, we collect the optimal action given by the policy oracle. Then the policy is updated for 10 epochs over all the state-action pairs collected so far, using the RMSProp optimizer \citep{tieleman2012lecture}. Both methods use Huber loss \citep{huber1964robust} between the estimated policy and the optimal policy given by the policy oracle.

For reinforcement learning, we run experiments with A3C algorithm. 
The policy learning module has a linear layer of size 256 followed by an LSTM layer of size 256 which encodes the history of state observations. The LSTM layer's output is fully-connected to a single neuron to predict the value function as well as three other neurons to predict the policy function. 
All the network parameters are shared for predicting both the value function and the policy function except the final fully connected layer. 
All the convolutional layers and fully-connected linear layers have ReLu activations \citep{nair2010rectified}.
The A3C model was trained using Stochastic Gradient Descent (SGD) with a learning rate of 0.001. We used a discount factor of 0.99 for calculating expected rewards and run 16 parallel threads for each experiment. 
We use mean-squared loss between the estimated value function and discounted sum of rewards for training with respect to the value function, and the policy gradient loss using for training with respect to the policy function.

\section{Results \& Discussions}
\label{sec:res}
%\vspace{-0.05in}
For all the models described in section \ref{sec:model-arch}, the performance on both Multitask and Zero-shot Generalization is shown in Table~\ref{tab:results}. The performance of A3C models on Multitask Generalization during training is plotted in Figure \ref{fig:policy_evaluation}. 

\textbf{Performance of GA models:} We observe that models with the Gated-Attention (GA) unit outperform models with the Concatenation unit for Multitask and Zero-Shot Generalization. From Figure~\ref{fig:policy_evaluation} we observe that A3C models with GA units learn faster than Concat models and converge to higher levels of accuracy. In hard mode, GA achieves $83\%$ accuracy on Multitask Generalization and $73\%$ on Zero-Shot Generalization, whereas Concat achieves $24\%$ and $12\%$ respectively and fails to show any considerable performance. For Imitation Learning, we observe that GA models perform better than Concat, and that as the environment modes get harder, imitation learning does not perform very well as there is a need for exploration in medium and hard settings.
In contrast, the inherent extensive exploration of the reinforcement learning algorithm makes the A3C model more robust to the agent's location and covers more state trajectories. 

\begin{figure*}
\includegraphics[width=\linewidth,height=\textheight,keepaspectratio]{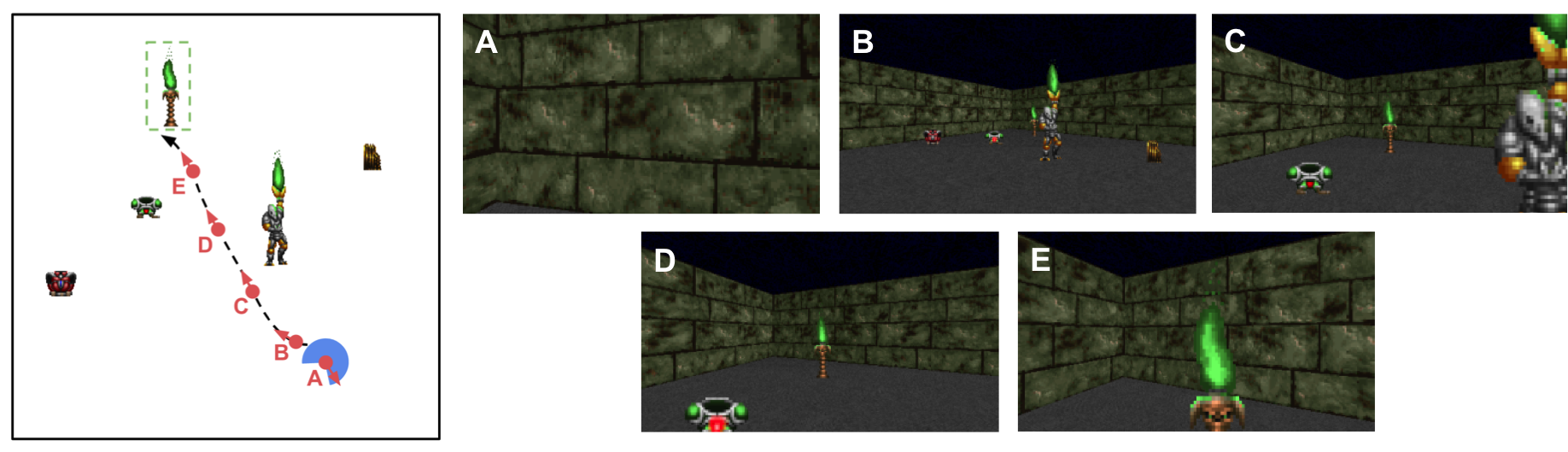}
%\vspace{-24pt}
\caption{\small This figure shows an example of the A3C policy execution at different points for the instruction `Go to the short green torch'. \textit{Left}: Navigation map of the agent, \textit{Right}: frames at each point. \textbf{A:} Initial frame: None of the objects are visible. \textbf{B:} agent has turned so that objects are in the field of view. \textbf{C : } agent successfully avoids the tall green torch. \textbf{D :} agent moves towards the short green torch. \textbf{E :}agent reaches target.}
\label{fig:policy_example}
%\vspace{-5pt}
\end{figure*}

\textbf{Policy Execution} : Figure \ref{fig:policy_example} shows a policy execution of the A3C model in the hard mode for the instruction \textit{short green torch}. In this figure, we demonstrate the agent's ability to explore the environment and handle occlusion. In this example, none of the objects are in the field-of-view of the agent in the initial frame.
The agent explores the environment (makes a ~300 degree turn) and eventually navigates towards the target object. It has also learned to distinguish between a \emph{short green torch} and \emph{tall green torch} and to avoid the tall torch before reaching the short torch\footnote{Demo videos: \url{https://goo.gl/rPWlMy}}.

\begin{figure*}
\minipage{0.72\textwidth}
\includegraphics[width=0.99\linewidth,keepaspectratio]{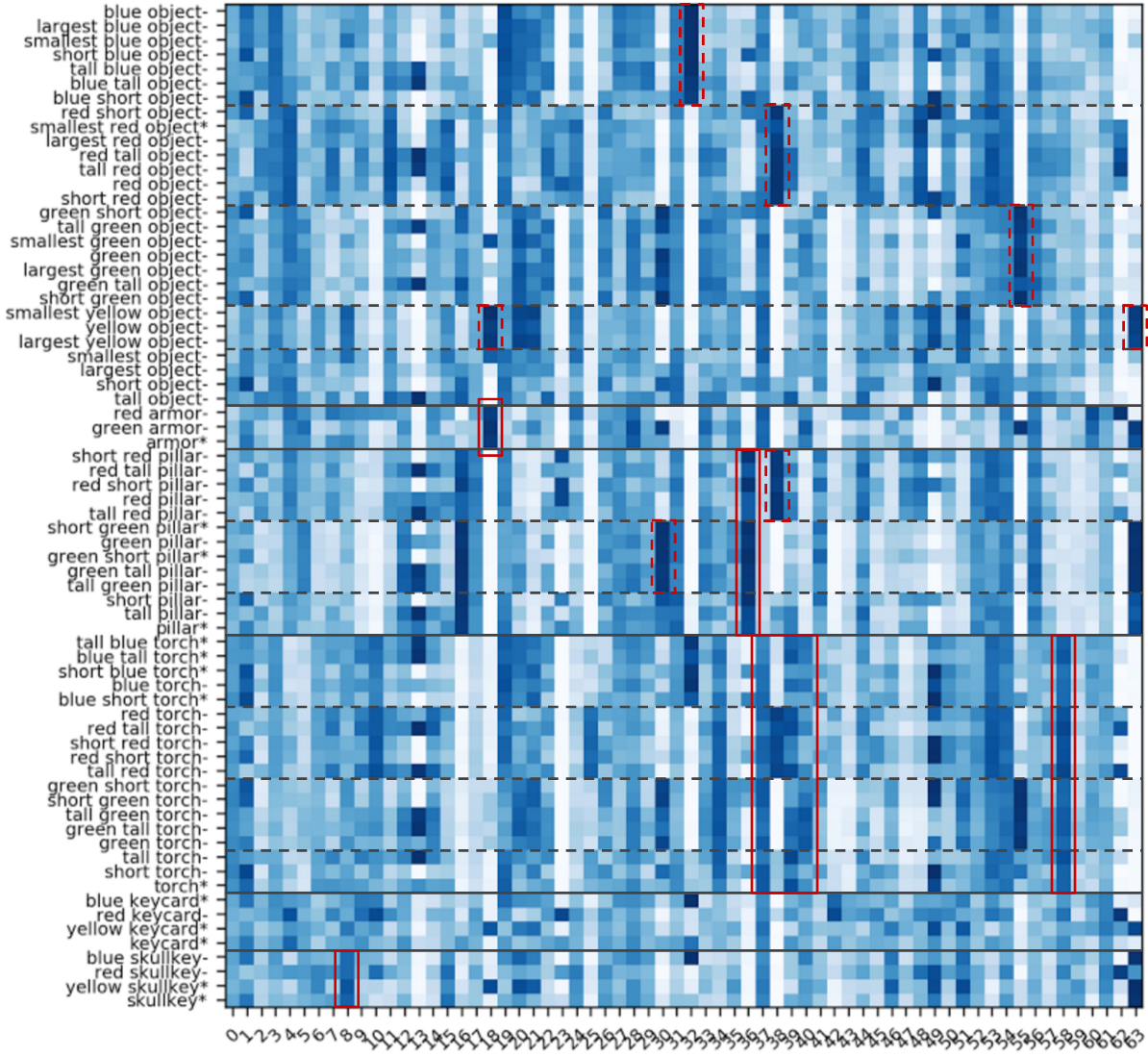}
\caption{\small Heatmap of the values of the 64-dimensional attention vector for different instructions grouped by object type and sub-grouped by object color. The test instructions are marked by *. The red boxes indicate that certain dimensions of the attention vector get activated for particular attributes of the target object referred in the instruction.}
%\vspace{-5pt}
\label{fig:analysis1}
\endminipage\hfill
\minipage{0.26\textwidth}
\includegraphics[width=0.99\linewidth,height=\textheight,keepaspectratio]{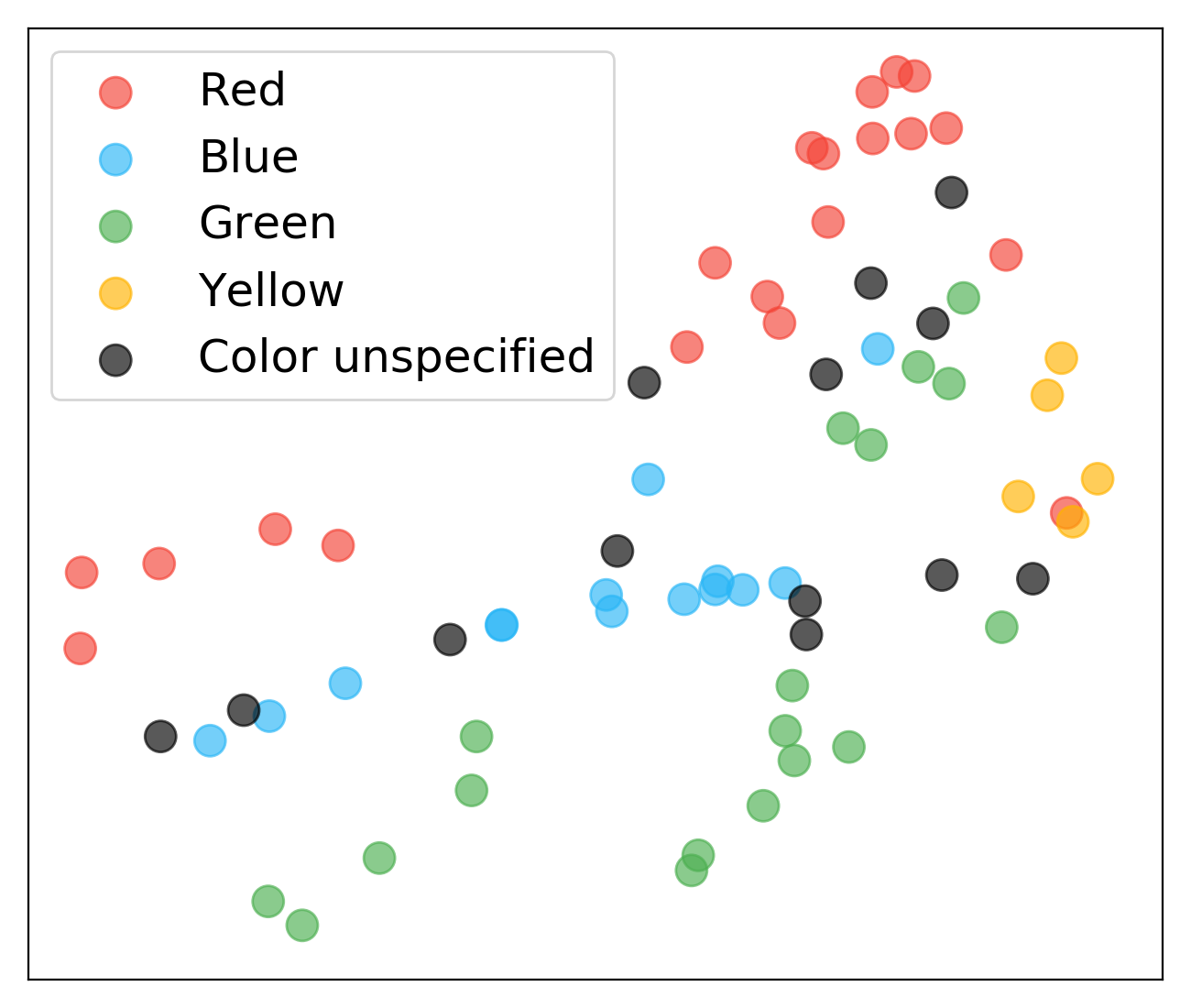}

\includegraphics[width=0.99\linewidth,height=\textheight,keepaspectratio]{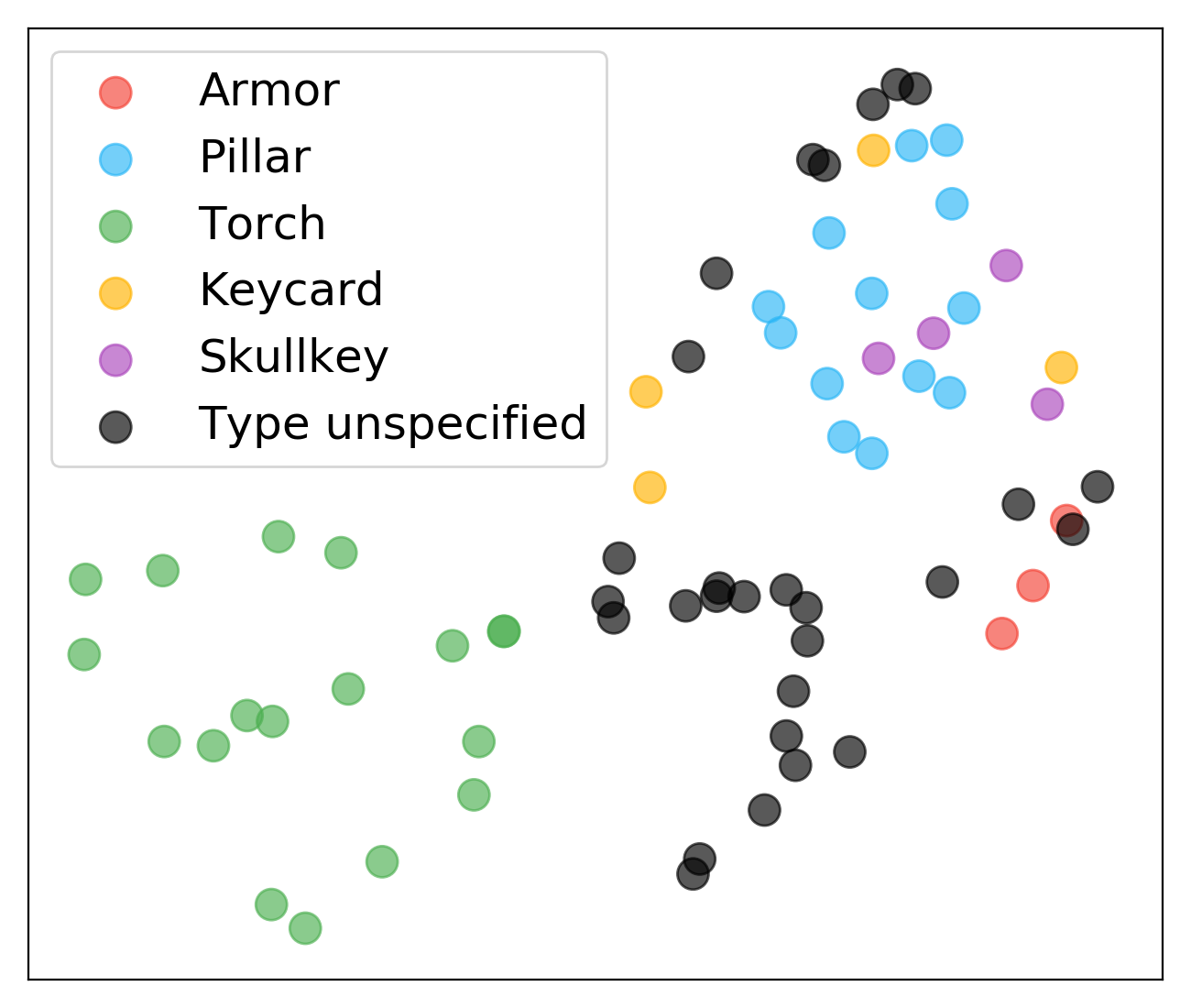}

\includegraphics[width=0.99\linewidth,height=\textheight,keepaspectratio]{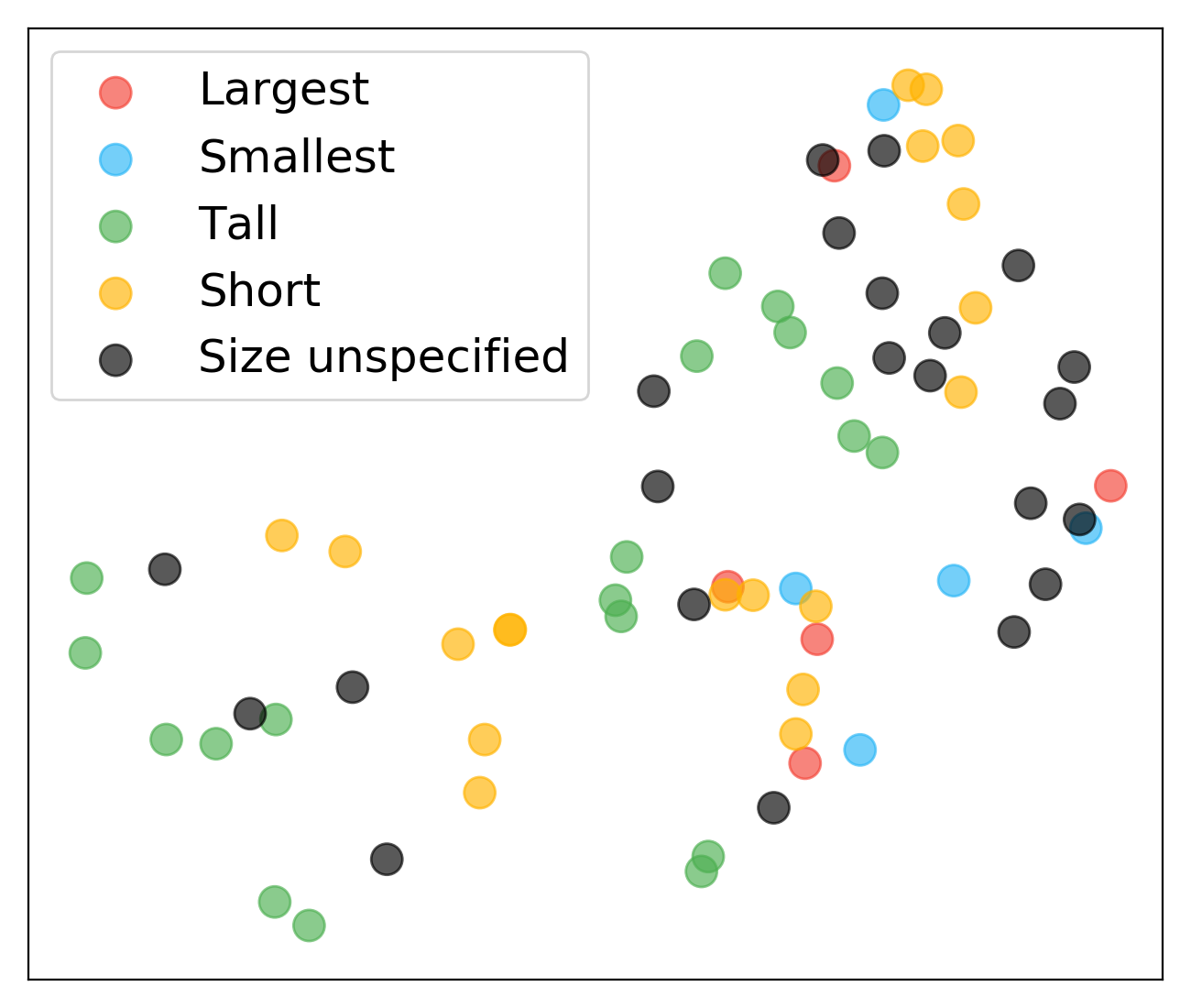}
	\caption{\small The t-SNE visualization of the attention vectors showing clusters based on object color, type and size.}
%\vspace{-5pt}
\label{fig:tsne}
\endminipage\hfill
%\vspace{-5pt}
\end{figure*}

\textbf{Analysis of Attention Maps:} Figure~\ref{fig:analysis1} shows the heatmap for values of the attention vector for different instructions grouped by object type of the target object (additional attention maps are given in the supplementary material).
As seen in the figure, dimension 18 corresponds to `armor', dimensions 8 corresponds to the `skullkey' and dimension 36 corresponds to the `pillar'. Also, note that there is no dimension which is high for all the instructions in the first group. This indicates that the model also recognizes that the word `object' does not correspond to a particular object type, but rather refers to any object of that color (indicated by dotted red boxes in \ref{fig:analysis1}). These observations indicate that the model is learning to recognize the attributes of objects such as color and type, and specific feature maps are gated based on these attributes. 
Furthermore, the attention vector weights on the test instructions (marked by * in figure ~\ref{fig:analysis1}) also indicate that the Gated-Attention unit is also able to recognize attributes of the object in unseen instructions.
We also visualize the t-SNE plots for the attention vectors based on attributes, color and object type as shown in Figure~\ref{fig:tsne}. 
The attention vectors for objects of red, blue, green, and yellow are present in clusters whereas those for instructions which do not mention the object's color are spread across and belong to the clusters corresponding to the object type. Similarly, objects of a particular type present themselves in clusters. 
The clusters indicate that the model is able to recognize object attributes as it learns similar attention vectors for objects with similar.

%%\vspace{-0.1in}
\section{Conclusion}
\label{sec:conc}
%\vspace{-0.05in}
In this paper we proposed an end-to-end architecture for task-oriented language grounding from raw pixels in a 3D environment, for both reinforcement learning and imitation learning.  
The architecture uses a novel multimodal fusion mechanism, \textit{Gated-Attention}, which learns a joint state representation based on multiplicative interactions between instruction and image representation. 
We observe that the models 
(A3C for reinforcement learning and Behavioral Cloning/DAgger for imitation learning) 
which use the Gated-Attention unit outperform the models with concatenation units for both Multitask and Zero-Shot task generalization, across three modes of difficulty. 
The visualization of the attention weights for the Gated-Attention unit indicates that the agent learns to recognize objects, color attributes and size attributes.

\section*{Acknowledgements}
We would like to thank Prof. Louis-Philippe Morency and Dr. Tadas Baltrušaitis for their valuable comments and guidance throughout the development of this work. 
This work was partially supported by BAE grants ADeLAIDE FA8750-16C-0130-001 and ConTAIN HR0011-16-C-0136.

{
\bibliography{aaai2018}
\bibliographystyle{aaai}
}
%\end{document}
\onecolumn
\appendix
\section{Doom objects}
\label{sec:objects}

\begin{figure}[h]
\centering
\includegraphics[width=0.6\linewidth,height=\textheight,keepaspectratio]{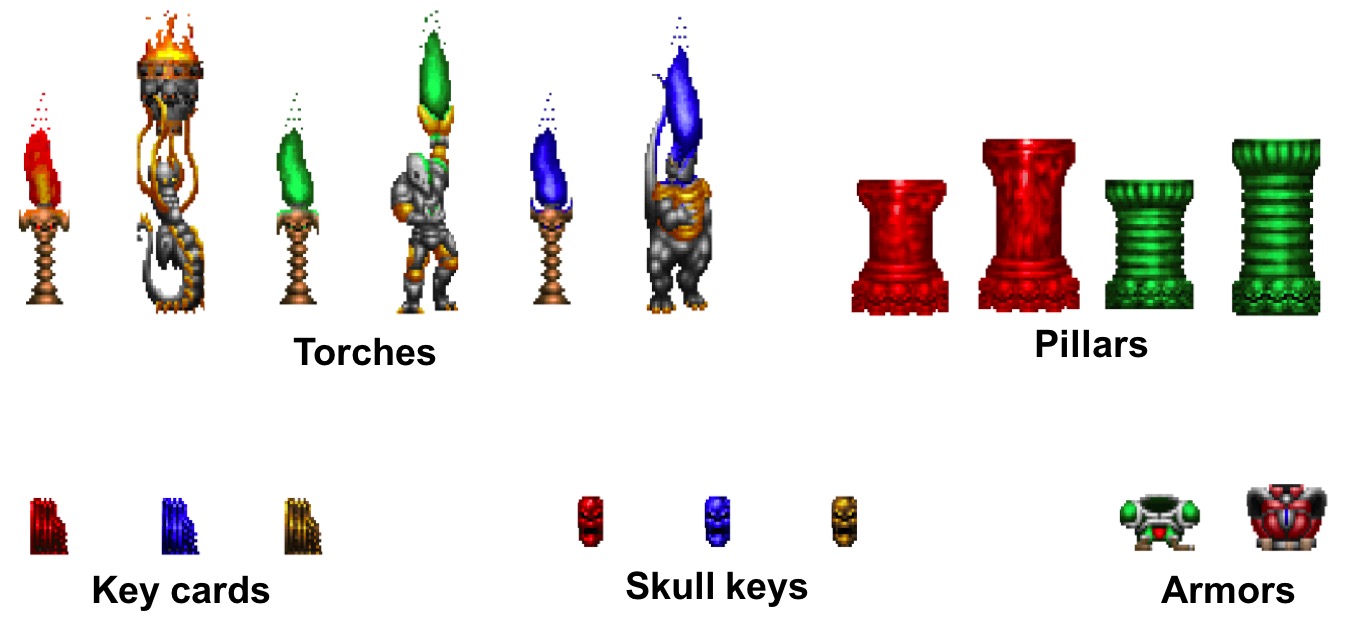}
\vspace{5pt}
\caption{\small Objects of various colors and sizes used in the environment}
  \label{fig:object-images}
\end{figure}

The ViZDoom environment supports spawning of several objects of various colors and sizes. The types of objects available are Columns, Torches, Armors and Keycards.  In our experiments, we use several of these objects, which are shown in Figure \ref{fig:object-images}. 

\section{Instructions}
\label{sec:instructions}

The list of 70 navigational instructions that was used to train and test the system in given in the Table \ref{tab:instructions}. 

\begin{table*}[h]
\centering
\begin{tabular}{@{}ll@{}}
\toprule
Instruction Type      & Instruction                                                                                                                                                                                                                                                                                                    \\ \midrule
Size + Color          & tall green torch, short red object, short red pillar, short red torch, tall red object,  \\ 
                            & tall blue object, tall green object, tall red pillar, tall green pillar, short blue torch, \\ 
                            & tall red torch, short green torch, short green object, short blue object, \\ 
                            & tall blue torch, short green pillar               \\ \midrule
Color + Size          & red short object, green tall torch, red short pillar, red short torch, red tall object,\\ 
							 & green tall object, blue tall object, red tall pillar,  green tall pillar,  \\
							 & red tall torch, blue tall torch, green short object, green short torch, \\ 
							 & blue short object, green short pillar, blue short torch \\ \midrule
Color                   & blue torch, red torch, green torch, yellow object, \\ 
							 & green armor, tall object, red skullkey, red object, green object \\
							 & blue object, red pillar, green pillar, red keycard, red armor, blue skullkey, \\
							 & blue keycard, yellow keycard, yellow skullkey  \\ \midrule
Object Type         & torch, keycard, skullkey, pillar, armor   \\ \midrule
SuperlativeSize+Color  & smallest yellow object, smallest blue object, smallest green object, \\
									& largest blue object, largest red object, largest green object, \\
									& largest yellow object, smallest red object \\ \midrule
SuperlativeSize     &   largest object, smallest object      \\ \midrule
Size                  & short torch, tall torch ,tall pillar ,short pillar ,short object, tall object                                                                                                                                                                                                                                   \\ \bottomrule
\end{tabular}
\caption{List of instructions. Each instruction of \emph{Go to the X}, where each `X' is each entry in the table}
\label{tab:instructions}
\end{table*}

\section{Attention Maps}
\label{sec:attention_maps}

\begin{figure*}
\centering
\includegraphics[width=0.68\linewidth,height=\textheight,keepaspectratio]{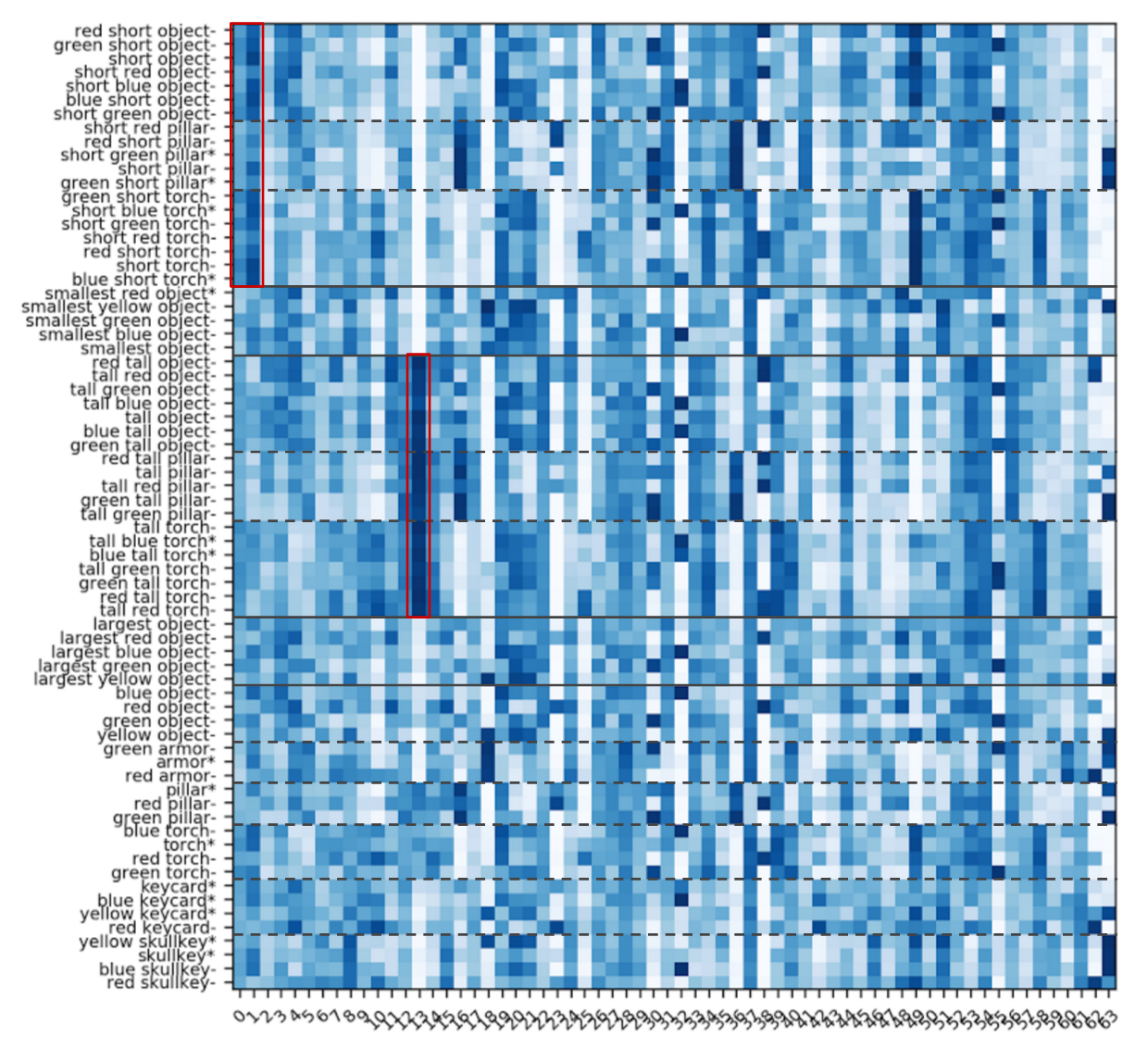}
%\vspace{5pt}
\caption{Attention vector output for different instructions grouped by description. The test instructions are marked by *.}
\label{fig:objects_1}
\end{figure*}

\begin{figure*}
\centering
\includegraphics[width=0.68\linewidth,height=\textheight,keepaspectratio]{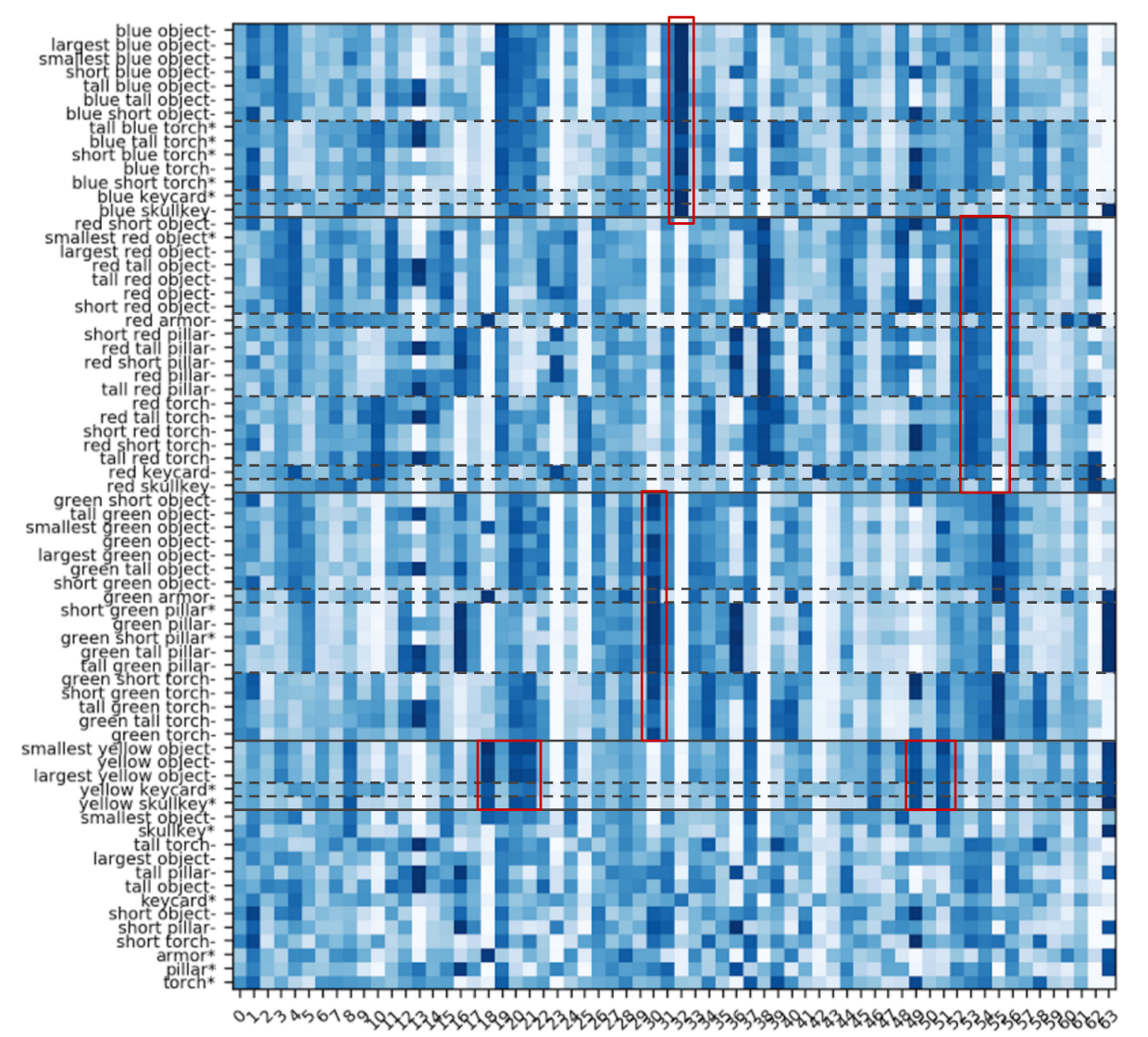}
%\vspace{5pt}
\caption{Attention vector output for different instructions grouped by color. The test instructions are marked by *.}
\label{fig:objects_2}
\end{figure*}

The attention maps for different instructions grouped based on description is shown in \ref{fig:objects_1} and grouped based on color is shown in figure \ref{fig:objects_2}.

\end{document}